# Recognition of Text Image Using Multilayer Perceptron

Vijendra Singh, Hemjyotsana Parashar and Nisha Vasudeva

*Faculty of Engineering & Technology,*
*Mody Institute of Technology & Science,*
*Lakshmangarh, Sikar, Rajasthan, India*
d_vijendrasingh@yahoo.co.in and
vasudeva.nisha1@gmail.com

*Abstract*——**The biggest challenge in the field of image processing is to recognize documents both in printed and handwritten format. Optical Character Recognition (OCR) is a type of document image analysis where scanned digital image that contains either machine printed or handwritten script input into an OCR software engine and translating it into an editable machine readable digital text format. A Neural network is designed to model the way in which the brain performs a particular task or function of interest: The neural network is simulated in software on a digital computer. Character Recognition refers to the process of converting printed Text documents into translated Unicode Text. The printed documents available in the form of books, papers, magazines, etc. are scanned using standard scanners which produce an image of the scanned document. Lines are identifying by an algorithm where we identify top and bottom of line. Then in each line character boundaries are calculated by an algorithm then using these calculation, characters is isolated from the image and then we classify each character by basic back propagation. Each image character is comprised of 30×20 pixels. We have used the Back propagation Neural Network for efficient recognition where the errors were corrected through back propagation and rectified neuron values were transmitted by feed-forward method in the neural network of multiple layers**

*Keywords*- **Back Propagation Algorithm, Character Recognition, Multi-Layer Perceptron, Supervised Learning**

## I. INTRODUCTION

Character recognition (in general, pattern recognition) addresses the problem of classifying input data, represented as vectors, into categories. Character Recognition is a part of Pattern Recognition [1].It is impossible to achieve 100% accuracy. The most basic way to recognizing the patterns using probabilistic methods in which [2] we use Bayesian Network classifiers for recognizing characters. The need for character recognition software has increased much since the outstanding growth of the Internet. Optical Character Recognition (OCR) is a very well-studied problem in the vast area of pattern recognition. Its origins can be found as early as 1870 when an image transmission system was invented which used an array of photocells to recognize patterns. Until the middle of the 20th century OCR was primarily developed as an aid to the visually handicapped. With the advent of digital computers in the 1940s, OCR was realized as a data processing approach for the first time. The first commercial OCR systems began to appear in the early 1950s and soon they were being used by the US postal service to sort mail. The accurate recognition of Latin-script, typewritten text is now considered largely a solved problem on applications where clear imaging is available such as scanning of printed documents [3, 4]. Typical accuracy rates on these exceed 99%. Total accuracy can only be achieved by human review. Optical Character Recognition (OCR) programs are capable of reading printed text. This could be text that was scanned in form a document, or hand written text that was drawn to a hand-held device, such as a Personal Digital Assistant (PDA). OCR programs are used widely in many industries. Many of today's document scanners for the PC come with OCR software that allows you to scan in a printed document and then convert the scanned image into an electronic text format such as a word document, enabling you to manipulate the text. In order to perform this conversion the software must analyze each group of pixels (0's and 1's) that form a letter and produce a value that corresponds to that letter. Some [5] of the OCR software on the market uses a neural network as the classification engine. The original document is scanned into the computer and saved as an image. The character recognition software breaks the image into sub-images, each containing a single character. The sub-images are then translated from an image format into a binary format, where each 0 and 1 represents an individual pixel of the sub image. The binary data is then fed into a neural network that has been trained to make the association between the character image data and a numeric value that corresponds to the character. The output from the neural network is then translated into ASCII text and saved as a file. Recognition of characters is a very complex problem.

The characters could be written in different size, orientation, thickness, format and dimension. This will give infinite variations. The capability of neural network to generalize and insensitive to the [6, 7] missing data would be very beneficial in recognizing characters. The goal of this paper is to create an application interface for Character recognition that would use an Artificial Neural Network as the backend to solve the recognition problem. Neural Network used for training of neural network. The main driving force behind neural network research is the desire to create a machine that works similar to the manner our own brain works. Neural networks have been used in a variety of different areas to solve a wide range of problems. Unlike human brains that can identify and memorize the characters like letters or digits; computers treat them as binary graphics. Therefore, algorithms are necessary to identify and recognize each character. The features of directions of pixels of the characters with respect to their neighboring pixels are extracted and given as input to the neural network. Self [9] adaptive training algorithm is used for identification of variable learning rate. In k-nearest neighbor (k-NN) algorithm, the pattern class is obtained by looking into k number of nearest pattern sets that have the least Euclidean distance with that pattern itself. Support vector machine (SVM) and SOM is also used for train the neural network. In this paper offline recognition of character is done for this printed text document is used. It is a process by which we convert printed document or scanned page to ASCII character that a computer can recognize. A back propagation feed-forward neural network is used to recognize the characters. After training the network with back-propagation learning algorithm, high recognition accuracy can be achieved. Recognition of printed characters is itself a challenging problem since there is a variation of the same character due to change of fonts or introduction of different types of noises. Difference in font and sizes makes recognition task difficult if preprocessing, feature extraction and recognition are not robust.

This paper is organized as follows. Multilayer Perceptron Neural Network for Recognition is briefly described in section II. In section III, training performance and accuracy of prediction is analyzed. Section IV contains data description and result analysis. Finally, we conclude in Section V.

## II. Designing Of Multilayer Perceptron Neural Network For Recognition

There are two basic methods used for OCR: Matrix matching and feature extraction. Of the two ways to recognize characters, matrix matching is the simpler and more common. But still we have used Feature Extraction to make the product more robust and accurate. Feature Extraction is much more versatile than matrix matching.

Here we use Matrix matching for Recognition of character. The Process of Character Recognition of the document image mainly involves six phases:
- Acquisition of Grayscale Image
- Digitization/Binarization
- Line and Boundary Detection
- Feature Extraction
- Feed Forward Artificial Neural Network based Matching.
- Recognition of Character based on matching score.

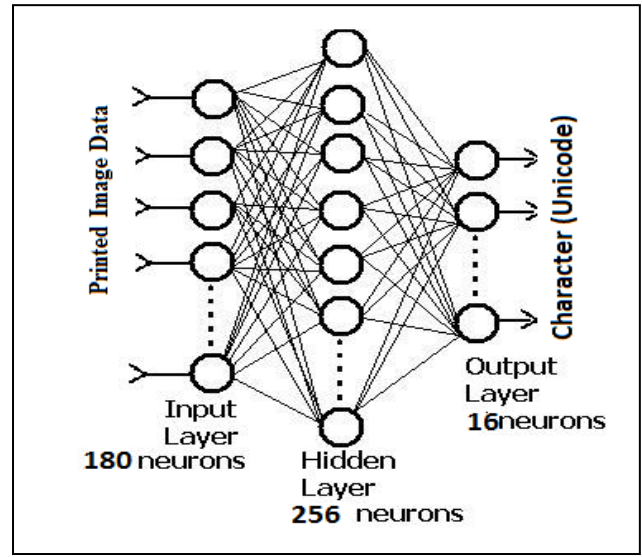

Fig 1. Multilayer Perceptron Neural Network

The multilayer Perceptron neural networks with the EBP algorithm have been applied to the wide variety of problems the acquisition phase uses a scanner or digital camera that catches photocopy of the text document as an image. The scanned image must be [4, 5] a grayscale image or binary image, where binary image is a contrast stretched grayscale image. That grayscale image is then undergoes digitization. In digitization [12] a rectangular matrix of 0s and 1s are formed from the image. Where 0-black and 1-white and all RGB values are converged into 0s and 1s. The matrix of dots represents two dimensional arrays of bits. Digitization is also called binarization as it converts grayscale image into binary image using adaptive threshold. Line and Boundary detection is the process of identifying points in a digital image at which the character top, bottom, left and right are calculated. Feed Forward Neural Network approach is used to combine all the unique features, which are taken as inputs, one hidden layer is used to integrate and collaborate[9] similar features and if required adjust the inputs by adding or subtracting weight values, finally one output layer is used to find the overall matching score of the network.

## III. TRAINING ALGORITHM PERFORMANCE AND ACCURACY OF PRIDICTION

The back propagation algorithm requires a numerical representation for the characters. Learning is implemented using the back-propagation algorithm with learning rate. Gradient is calculated [10] after every iteration and compared with threshold gradient value. If gradient is greater than the threshold value then it performs next iteration. The batch steepest descent training function is trained. The weights and biases are updated in the direction of the negative gradient of the performance function.

In order to determine quantitatively the model, two error measures is employed for evaluation and model comparison, being these: The model squared error (MSE) and the mean absolute error (MAE). If $y_i$ is the actual observation for a time period t and $F_t$ is the forecast for the same period, then the error is defined as

$$E_i = y_t - F_t \quad (1)$$

The standard statistical error measures can be defined as

$$MSE = \frac{1}{n} \sum_{i=1}^{n} e_{i=1}^{n} \quad (2)$$

And the mean absolute error as

$$MSE = \frac{1}{n} \sum_{i=1}^{n} |e_i| \quad (3)$$

Where n is the *n*umber of periods of time. When the mean square error decreased gradually and became stable, and the training and testing error produced satisfactory results .the training performance curve of neural network. The accuracy of the trained network is tested against output data. The accuracy of the trained network is assessed in the following way: in first way, the predicted output value is compared with the measured values .The results are presented shows the relative accuracy of the predicted output. The overall percentage error obtained from the tested results is 4%. In the second way, the root mean square error and the mean absolute error are determined and compared. The performance index for training of ANN is given in terms of mean square error (MSE).The tolerance limit for the MSE is set to 0.001.The MSE of the training set become stable at 0.0070 when the number of iteration reaches 350. The closeness of the training and the testing errors validates the accuracy of the model.

## IV. EXPERIMENTAL RESULTS

We developed our proposed system on the computer of 256 RAM, 250GB Hard Disk and windows 7 operating system. We create interface for proposed system for character recognition by using Microsoft Visual C # 2008 Express Editions.

The MLP network that is implemented is composed of three layers input layer, output layer and hidden layer. The input layer constitutes of 180 neurons which receive printed image data from a 30x20 symbol pixel matrix. The hidden layer constitutes of 256 neurons whose [12] number is decided on the basis of optimal results on a trial and error basis. The output layer is composed of 16 neurons.

Data Set:-we apply our system with different epochs values and no of neurons in hidden layer. We use different font style of printed text like Arial, Tahoma, Bookman old style and Times New Roman.The results of traning and testing phases are shown in fig 2 and fig 3.

Number of characters=90, Learning rate=150, No of neurons in hidden layer=256

**Table 1 Percentage of Error for different epochs**

| No of epochs / Font Style | 300 | 600 | 900 |
|---|---|---|---|
| | Error | | |
| Arial | 3.44% | 2.33% | 1.11% |
| Tahoma | 2.11% | 1.11% | 0 |
| Times new Roman | 0 | 0 | 1.11% |
| Bookman old style | 2.11% | 1.11% | 0 |

Number of characters=90, Learning rate=150, No of epochs=900.

**Table 2 Percentage of Error for different numbers of neurons**

| No of neurons in Hidden Layers / Font Style | 64 | 128 | 256 |
|---|---|---|---|
| | Error | | |
| Arial | 6.25% | 2.33% | 1.11% |
| Tahoma | 1.11% | 0 | 0 |
| Times new Roman | 2.33% | 0 | 1.11% |
| Bookman old style | 4.64% | 2.24% | 0 |

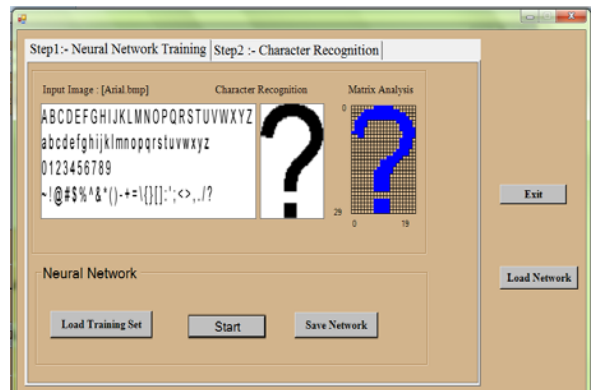

Fig 2 Training Phase

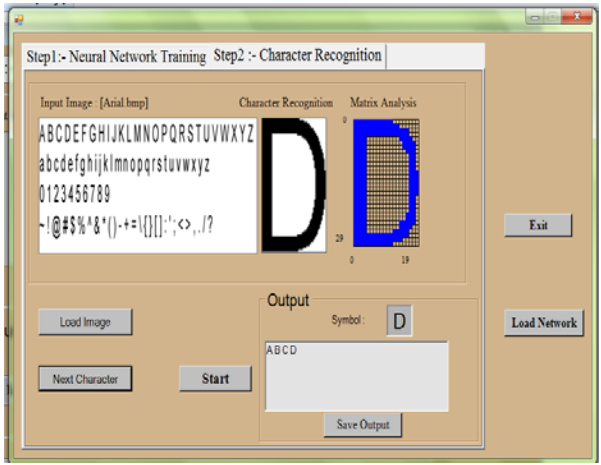

Fig 3 Testing Phase

## V. CONCLUSION

The important feature of this ANN training is that the learning rates are dynamically computed each epoch by an interpolation map. The ANN error function is transformed into a lower dimensional error space and the reduced error function is employed to identify the variable learning rates. As the training progresses the geometry of the ANN error function constantly changes and therefore the interpolation map always identifies variable learning rates that gradually reduce to a lower magnitude. As a result the error function also reduces to a smaller terminal function value. The result of structure analysis shows that if the number of hidden nodes increases the number of epochs taken to recognize the handwritten character is also increases. A lot of efforts have been made to get higher accuracy but still there are tremendous scopes of improving recognition accuracy

REFERENCES

[1] Andrew Blais and David Mertz, "An Introduction to Neural Networks Pattern Learning with Back Propagation Algorithm", Gnosis Software, Inc., July 2001
[2] David Bouchain,"Character Recognition Using Convolutional Neural Networks", Seminar Statistical Learning Theory University of Ulm, Germany Institute for Neural Information Processing Winter 2006/2007
[3] Yuelong Li Jinping Li Li Meng, "Character Recognition Based on Hierarchical RBF Neural Networks" *Intelligent Systems Design and Applications*, 2006. ISDA '06. Sixth International Conference, **1**, On Page(s): 127-132, 2006.
[4] Dong Xiao Ni Seidenberg," Application of Neural Networks to Character Recognition", CSIS, Pace University, May 4th, 2007, School of CSIS, Pace University, White Plains, NY.
[5] Sutha.J, Ramraj.N, "Neural Network Based Offline Tamil Handwritten Character Recognition System", *IEEE International Conference on Computational Intelligence and Multimedia Application*, 2007, **2**, 13-15, Dec.2007, Page(s),446-450, 2007.
[6] Velappa Ganapathy, and Kok Leong Liew ,"Handwritten Character Recognition Using Multiscale Neural Network Training Technique", World Academy of Science, Engineering and Technology, 2008
[7] Dayashankar Singh, Maitreyee Dutta and Sarvpal H. Singh, "Neural Network Based Handwritten Hindi Character Recognition", *ACM International Conference (Compute 09)*, Jan. 9-10, 2009, Bangalore.
[8] Hongxi Wei Guanglai Gao, "Machine-Printed Traditional Mongolian Characters Recognition Using BP Neural Networks ",Sch. of Comput. Sci., Inner Mongolia Univ., Hohhot, China,1 - 7,Dec. 2009
[9] Anita Pal1 and Dayashankar Singh, Handwritten English Character Recognition Using Neural Network",Department of Computer Science & Engineering, U.P.Technical University, Lucknow, India International Journal of Computer Science & CommunicationVol. 1, No. 2, July-December 2010, pp. 141-144
[10] Omaima N.A. AL-Allaf, "Improving the Performance of Backpropagation Neural Network Algorithm for Image Compression/Decompression System",Department of Computer Information Systems,Faculty of Sciences and Information Technology, AL-Zaytoonah Private University of Jordan, P.O. Box 130, Amman (11733), Jordan Journal of Computer Science 6 (11): 1347-1354, 2010
[11] Raghuraj Singh, C. S. Yadav, Prabhat Verma, Vibhash Yadav,"Optical Character Recognition (OCR) for Printed Devnagari Script Using Artificial Neural Network",International Journal of Computer Science & Communication Vol. 1, No. 1, 2010, pages 91-95.
[12] simon S. Haykin " Neural Networks A Comprehensive Foundation", prentice hall , chapter 1-15 , 1990,page 1-889.
[13] Christopher M.Bishop"Neural Networks For Pattern Recognition" Department of computer Science and applied mathematics Aston University , Birmingham, UK Clarndon press, 1995.